# Learning Bayesian Networks with Restricted Causal Interactions

**Julian R. Neil, Chris S. Wallace** and **Kevin B. Korb**
School of Computer Science and Software Engineering
Monash University, Clayton, Vic., 3168, Australia

## Abstract

A major problem for the learning of Bayesian networks (BNs) is the exponential number of parameters needed for conditional probability tables. Recent research reduces this complexity by modeling local structure in the probability tables. We examine the use of log-linear local models. While log-linear models in this context are not new (Whittaker, 1990; Buntine, 1991; Neal, 1992; Heckerman and Meek, 1997), it is generally subsumed under a naive Bayes model. We describe an alternative using a Minimum Message Length (MML) (Wallace and Freeman, 1987) metric for the selection of local models with causal independence, which we term a *first-order model* (FOM). We also combine FOMs and full conditional models on a node-by-node basis.

## 1 INTRODUCTION

While methods for learning Bayesian networks (BNs) and probabilistic inference using them continue to improve, their application is limited by the fact that both problems are NP-hard (Cooper, 1990; Chickering et al., 1994). One factor contributing to this complexity is the exponential number of parameters needed to specify fully the conditional probability tables (CPTs). Recent research has examined the prospect of reducing the number of parameters by modeling local structure in the probability tables. Such a reduction in BN complexity broadens the range of problems accessible to modeling with BNs. In addition, if the restricted parameter set can still accurately represent the underlying causal interactions, estimating fewer parameters benefits discovery by allowing network structure and parameters to be inferred from smaller datasets than otherwise. While these ideas are not new (Pearl, 1988; Buntine, 1991), until recently there have been few experimental results supporting them.

Here we examine the use of local models in learning BNs and in particular log-linear models, which have been used extensively in the social sciences to investigate interactions in contingency tables. While this use of log-linear models has been suggested previously, this has generally been in the context of parameter rather than structure learning (Whittaker, 1990; Spiegelhalter and Lauritzen, 1990; Neal, 1992; Jordan et al., 1998). Where structure learning is mentioned, the log-linear model is commonly associated with a naive-Bayes approach (Buntine, 1991; Heckerman and Meek, 1997). We show that this is unnecessary and describe a restricted logit model exhibiting causal independence. We describe how Minimum Message Length (MML) (Wallace and Freeman, 1987) can be used to select between local CPT models as well as for discovering network structure. We give MML metrics for estimating the posterior probability of nodes in a BN for both a restricted logit model and the traditional full CPT. Finally, we choose several real datasets and use an MML Metropolis sampling algorithm to identify structures of high posterior probability, comparing networks using a logit CPT model with standard Bayesian networks and with a dual network that uses MML to select the CPT model on a none-by-node basis.

## 2 BAYESIAN NETWORKS

Bayesian networks model the joint distribution of the variables $\mathbf{V} = \{X_1, ... X_m\}$ by repeated conditionalization. Given a total ordering $\langle X_1, ..., X_m \rangle$, this results in the following decomposition:

$$P(\mathbf{V}) = \prod_{X_i \in \mathbf{V}} P(X_i | \mathbf{\Pi}_i) \qquad (1)$$

where $\mathbf{\Pi}_i \subseteq \{X_1, ..., X_{i-1}\}$. This is readily represented by directed acyclic graphs (dags) where nodes in the graph have a one-to-one correspondence with variables in $\mathbf{V}$ and edges indicate direct causal influence between variables. $\mathbf{\Pi}_i$ is then the set of parents of the node corresponding to variable $X_i$ in the dag. Any total ordering of the nodes in the graph in which no arc is directed towards an earlier node in the ordering — termed a *linear extension* of the dag — satisfies (1). The causal interactions between $\mathbf{\Pi}_i$ and $X_i$ are completely determined by the CPT $P(X_i|\mathbf{\Pi}_i)$.

An important property of BNs arises from the conditional independence relations that are implied by a model's structure. Assuming that our dag models satisfy the Markov property — i.e., every conditional independency implied by the model holds in the sampled population — it is possible for different dags to entail the same set of conditional independencies, and thus be capable of representing identical joint distributions over $\mathbf{V}$. Accordingly, such models are termed *statistically equivalent*. Two dags are statistically equivalent just in case they have identical undirected adjacencies and have identical *v-structures* — triples $\{X, W, Y\}$, where $X$ and $Y$ are not adjacent, but are both parents of $W$ — and therefore differ only in the orientation of one or more arcs. In contrast, a causal interpretation necessarily distinguishes any two models that differ even just in the orientation of a single arc, say $X \to Y$ and $Y \to X$. In one model $X$ is a cause of $Y$, while in the other $Y$ is a cause of $X$. To distinguish between two equivalent causal networks we must then rely on considerations other than observational data.

## 3  LOCAL PROBABILITY MODELS

In an effort to reduce the size of CPTs, one local probability model assumes that causes of a common effect do not interact with one another (e.g., noisy-or networks (Pearl, 1988; Dìez, 1993)). This effectively excludes the representation of non-monotonic causal interactions, and has been termed *causal independence* (Heckerman, 1993). Mitigating the reduction in the family of representable distributions is the fact that a real-valued causal independence model — linear models — have successfully been used for causal modeling in the social sciences for over five decades (Wright, 1934). This restrictive assumption has the advantage of reducing the parameter set for each variable to be linear in the number of parents.

More recently, research has concentrated on functional decompositions of the conditional distribution (Zhang and Poole, 1994; Meek and Heckerman, 1997) and specifically on the use of classification trees (Boutlier et al., 1996; Friedman and Goldszmidt, 1996), graphs (Chickering et al., 1997) and rule sets (Poole, 1998). The use of these structures have a twofold benefit: first, there is a large volume of literature on their induction from data; and second, they provide a richer causal interaction semantics.

Surprisingly, there has been little interest in the use log-linear local models, particularly for structure learning. Log-linear models are often used in the social sciences to represent models of causal interaction in contingency tables. There are many statistical tests (see Bishop et al., 1975; Goodman, 1978; Whittaker, 1990) and Bayesian methods (Darroch et al., 1980; Laskey and Martignon, 1996) for log-linear representations of contingency tables from data. Where log-linear models have been suggested for BN structure learning (Buntine, 1991; Heckerman and Meek, 1997) theoretical and experimental results have not been presented and generally the log-linear approach is placed in the context of a *naive*-Bayes model (see §3.2 below). Comparative structure learning results with the traditional BN are unavailable.

### 3.1  THE LOGIT MODEL

For a single variable $Y \in \mathbf{V}$ and its set of direct causes $\mathbf{\Pi}$, the traditional full conditional probability model gives the following probability parameterization:

$$P(Y = y|\mathbf{\Pi} = \pi_j, \xi) = \theta_{y\pi_j} \qquad (2)$$

where $\sum_y \theta_{y\pi_j} = 1$, $\pi_j$ is the $j^{\text{th}}$ instantiation of $\mathbf{\Pi}$ and $\xi$ describes our prior knowledge including the joint distribution $P(\mathbf{\Pi})$. We use $\Theta = \{\theta_{y\pi_j}\}$ to represent the complete parameter set and $\Theta'$ to be the set of free parameters. For notational simplicity we drop $\xi$ from further equations. Although capable of expressing the complete set of joint distributions over i.i.d. data, this parameterization does not explicitly represent causal independencies in the interactions between $\mathbf{\Pi}$ and $Y$. An equally rich log-linear representation known as a *logit* model makes such independencies inherent in the parameterization. For the sake of simplicity we initially consider a binary effect $Y$ with values $y_1$ and $y_0$ and two binary causes $W$ and $Z$. We further let $P(y|w, z) \equiv P(Y = y|W = w, Z = z)$. The logit model then uses the following parameterization:

$$\log \frac{P(y_1|w, z)}{P(y_0|w, z)} = a + b\text{T}(w) + c\text{T}(z) + d\text{T}(w.z) \qquad (3)$$

where $\text{T}(\cdot) = 1$ if its argument is true and 0 if its argument is false and $w.z$ is the conjunction of $w$ and $z$. It is evident from this model that $a$ is a parameter reflecting the propensity for $Y$ to be true independently of $W$ and $Z$, what we term a zeroth-order interaction, while $b$ gives the tendency for $Y$ to be true when $W$ is true, a first-order interaction. Similarly $c$ represents the association between $Y$ and $Z$. Lastly, $d$ can be seen to control the joint effect of $W$ and $Z$ on $Y$, and is thus a second-order interaction. Clearly increasing the number of parents of $Y$ requires higher-order parameters to represent the full conditional distribution.

The logit model has several advantages. First, when warranted, we can impose a causal interaction structure by setting some parameters to zero. This reduces the number of parameters, allowing accurate parameter estimation with smaller samples. Models with parameters removed are generally termed *unsaturated* *models*, and are considered *hierarchical* when fixing a parameter also fixes all of the higher-order parameters involving the same variables.

A second desirable property of the logit model is that it allows us to incorporate realistic prior beliefs about causal interaction structure. Parameters of the full conditional model $\Theta$ are generally assumed to be independent — i.e., given two parent states $\pi_1$ and $\pi_2$, the effect of altering the state of a single parent in $\pi_1$ is a priori assumed completely uncorrelated with the effect of identically altering the same parent in $\pi_2$. While such non-monotonic interactions do occur, the commoner pattern is that at least the direction of the effect of a change in one causal factor is not dependent on the values of other causal factors. A good model of discrete causal interactions should express the expectation that effects of one factor are commonly fairly consistent, and only mildly affected by other causal factors. It is straightforward to represent such expectations in the logit model by assuming a distribution over causal interaction structures. A simple example is a prior distribution that prefers hierarchical models of lower order. This expectation is difficult to define for the full conditional model of (2).

For causal discovery we would like to infer causal interaction structure from sample data. While there are statistical tests and Bayesian metrics for selecting between unsaturated log-linear models, searching the space of all possible unsaturated models is computationally expensive, particularly since we intend to search simultaneously for the overall network structure. Here we consider a first-order hierarchical model of the conditional distribution and assume interactions of second order and above are negligible. For the binary model of (3), this is equivalent to setting $d = 0$. Given the joint distribution $P(W, Z)$, this restriction can be shown to coincide with the maximum entropy distribution for $P(Y|W, Z)$ when $P(Y)$, $P(Y|W)$ and $P(Y|Z)$ are known, and thus can be said to restrict interactions to exhibit causal independence. The first-order model (FOM) has the same expressive power as the full conditional distribution when $Y$ has 0 or 1 parent, but with 2 or more parents the FOM can only accurately represent a subset of the distributions expressible by the full conditional model.[1]

Generalizing to non-binary variables, we replace the $T(\cdot)$ notation with subscripts and assume $Y$, $W$ and $Z$ are of arity $r_y$, $r_w$ and $r_z$ respectively. The first-order probability model becomes:

$$P(Y = k|w, z) = \frac{e^{a_k+b_{kw}+c_{kz}}}{\sum_{j=1}^{r_y} \left(e^{a_j+b_{jw}+c_{jz}}\right)} \quad (4)$$

where $a_k$ is the tendency for $Y = k$ independently of its parents, $b_{kw}$ reflects the tendency for $Y = k$ when $W = w$, and similarly for $c_{kz}$. It is apparent from (4) that this model is under-constrained.[2] We remove these degrees of freedom via the constraints:

$$\sum_k a_k = 0, \quad \sum_k b_{kw} = 0, \quad \sum_w b_{kw} = 0 \quad (5)$$

with corresponding constraints on $c_{kz}$. Although it is common to constrain parameters by arbitrarily setting several to zero — as we did in the binary case (3) — the constraints in (5) allow us to define a prior over the parameters that is invariant to relabelling the states of a variable. We refer to the full parameter set as $\Phi = \{c_k, a_{kw}, b_{kz}\}$ and the free parameters satisfying (5) as $\Phi'$. It is straightforward to generalize (4) to larger parent sets (Neil et al., 1999).

## 3.2  NAIVE-BAYES MODEL

The use of first-order log-linear or logit models in BNs is often associated with a naive-Bayes model (fig. 1). This model considers $X_1$, $X_2$ and $X_3$ to be conditionally independent given $Y$. Letting $\mathbf{X} = \{X_1, X_2, X_3\}$, this is equivalent to assuming the joint distribution:

$$P(Y, \mathbf{X}) = P(Y)P(X_1|Y)P(X_2|Y)P(X_3|Y) \quad (6)$$

with corresponding conditional probability parameter sets $\boldsymbol{\theta}_Y$, $\boldsymbol{\theta}_{X_1|Y}$, $\boldsymbol{\theta}_{X_2|Y}$ and $\boldsymbol{\theta}_{X_3|Y}$. These parameters can be inferred from data. While not a direct model of $Y$'s dependence on $\mathbf{X}$, this model can be transformed to give a conditional model:

$$P(Y, \mathbf{X}) = P(\mathbf{X})P(Y|\mathbf{X}) \quad (7)$$

Such a transformation results in a parameterization remarkably similar to (4). There is however one significant drawback to performing such a transformation. By necessity of the assumed joint distribution, the transformed distributions $P(\mathbf{X})$ and $P(Y|\mathbf{X})$ are dependent on one another, clearly a situation that is not intended by (7). Heckerman and Meek (1997) point out that for predicting $Y$ from $\mathbf{X}$, the use of any conditional model derived from an assumed joint distribution — other than one explicitly representing the requisite conditionalization — is suspect. This does not, however, give cause to discard the logit model of causal independence, nor does it confine its use in BNs to application of a naive-Bayes model. In contrast to the naive model we use the logit parameterization for the conditional distribution $P(Y|\mathbf{X})$ without placing any restrictions on $P(\mathbf{X})$. We do not assume a joint naive-Bayes model, but rather assume a maximum entropy distribution for P($Y|\mathbf{X}$) when P($\mathbf{X}$), P($Y$) and $\forall_{X_i \in \mathbf{X}}$P($Y|X_i$) are known. As such we are are estimating P($Y|\mathbf{X}$) assuming knowledge of the sampled distribution of P($\mathbf{X}$) rather that extracting it from an inferred joint model of P($Y, \mathbf{X}$).

---

[1] For binary variables a network using this conditional probability model is sometimes referred to as a *sigmoid belief network* (Neal, 1992).

[2] In particular, it is possible to add a constant to each $a_k : 1 \leq k \leq r_y$. Similarly, for $W = w$ we can translate each $b_{kw}$ (or $c_{kz}$ when $Z = z$) and leave the distribution unaltered. Finally, for $Y = k$, if we subtract a constant from $a_k$ and add the same value to each $b_{kw} : 1 \leq w \leq r_w$ (or $c_{kz} : 1 \leq z \leq r_z$), the conditional probabilities are left unchanged.

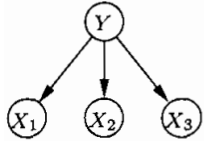

Figure 1: Naive-Bayes model.

## 4 MINIMUM MESSAGE LENGTH

Bayesian analysis assesses belief in a hypothesis $H$ by evaluating its posterior probability given sample data $D$, P($H|D$). If the task requires only the comparison of models then it suffices to calculate the joint probability P($H, D$) = $h(H)f(D|H)$ where $h(H)$ is our prior belief in $H$, $f(D|H)$ is the likelihood of obtaining the sample $D$ given $H$.

Full Bayesian analysis evaluates P($H, D$) by averaging over all possible parameterizations of $H$. In contrast, Wallace's Minimum Message Length Principle (MML) (Wallace and Freeman, 1987) considers parameter estimation to be an intrinsic part of inference and uses information theory to evaluate $-\log$P($H, D$) under an optimal partitioning of the parameter space.[3] The hypothesis and sample data are hypothetically encoded as a two part message, the first describing a parameterization of the hypothesis, and the second encoding the sample data under the assumption of the hypothesis. Choosing parameter estimates that minimize the total message length, hypotheses that result in shorter message lengths result in a tradeoff between hypothesis complexity and fit to the data.

For variable $Y$ with parents $\Pi$, our hypothesis $H$ is that the joint probability P($Y, \Pi$) = P($Y|\Pi$)P($\Pi$), where P($\Pi$) is known a priori. We consider two parameterizations, the full conditional model given by $\Theta$ in (2) and the FOM $\Phi$ of (4). Assuming i.i.d data, independent parameters and a uniform conjugate Dirichlet prior for $\Theta$, Cooper and Herskovits (1992) derive P($H, D$) for the full conditional model. Using the general result than an MML message exceeds the length of a non-estimating message — e.g., a full Bayesian posterior analysis — by about $\frac{1}{2}(\log \pi - \log 6 + 1)$ nits per parameter (Wallace and Freeman, 1987) we find the message length for transmitting $Y$ under the full conditional model is:

$$I_Y^{\Theta} = \frac{|\Theta'|}{2} \log \frac{\pi e}{6} - \sum_{\pi_l} \sum_k \log N_{k\pi_l}!$$
$$+ \sum_{\pi_l} [\log(N_{\pi_l} + r_y - 1)! - \log(r_y - 1)!] \quad (8)$$

where $|\Theta'|$ is the number of free parameters, $N_{k\pi_l}$ is the number of cases in $D$ where $Y = k$ and $\Pi = \pi_l$, and $N_{\pi_l} = \sum_k N_{k\pi_l}$.

We now describe a message length calculation for the FOM.[4] Wallace and Freeman (1987) derive an expression for evaluating message length of models satisfying broad regularity conditions. Their expression approximates the message length by optimally partitioning the parameter space to account for specification of the parameters with only limited accuracy:

$$I_{\text{WF}} = -\log \frac{Kh(\Phi')}{\sqrt{F(\Phi')}} - \log f(D|\Phi') + \frac{|\Phi'|}{2} \quad (9)$$

where $h(\Phi')$ is the parameter prior, $F(\Phi')$ is the expected Fisher Information, $f(D|\Phi')$ is the likelihood function and $K$ is a constant resulting from partitioning the parameter space. For the FOM, a choice for the parameter prior is not directly obvious. Although we would prefer a conjugate prior, it is unclear for variables of arbitrary arity how to define such a prior that is symmetric in its expectation over all parameters. We consider it a reasonable alternative to assume that the parameters are independently normally distributed with variance $\sigma^2$. Although this biases the conditional distributions away from extreme probabilities, choosing a suitably large variance reduces this bias. For this paper we set $\sigma = 3$. To maintain symmetry and invariance to relabelling the states of a variable, we assume an i.i.d. normal prior over all the parameters $\Phi$, rather than just the free parameters, and normalize the distribution to satisfy the constraints (5):

$$h(\Phi') = \frac{\sqrt{r_y} \prod_{X_i \in \Pi} \sqrt{r_y{}^{r_i - 1} r_i{}^{r_y - 1}}}{(\sqrt{2\pi}\sigma)^{|\Phi'|}} e^{-\frac{1}{2\sigma^2}\Phi^T \Phi} \quad (10)$$

Substituting into (9) and expanding the likelihood gives as an MML estimate for the first-order model:

$$I_Y^{\Phi} = \frac{d}{2}\log 2\pi + d\log \sigma - \frac{1}{2}\log r_y$$
$$- \frac{1}{2}\sum_{X_i \in \Pi}[(r_i - 1)\log r_y + (r_y - 1)\log r_i]$$
$$+ \frac{1}{2\sigma^2}\sum_{\phi \in \Phi}\phi^2 + \frac{1}{2}\log F(\Phi')$$
$$- \sum_{\pi_l}\sum_k N_{k\pi_l} \log P_{k\pi_l} + \frac{d}{2}(1 + \log \kappa_d) \quad (11)$$

---

[3] We use natural logarithms and report message lengths in "nits" rather than bits (1 nit = $1/\log_e 2$ bits).

[4] For a full derivation see Neil et al. (1999)

where $d = |\Phi'|$ is the number of free parameters and $\kappa_d$ is a lattice constant from partitioning the parameter space (Conway and Sloane, 1988). As it is difficult to derive a closed form expression for the Fisher Information for the FOM (Neil et al., 1999) it is not possible to define directly parameter estimates that minimize 11. Maximum likelihood (ML) estimates offer an alternative, as MML and ML estimates converge. But if some of the data counts $N_{y\pi_l}$ turn out to be zero, ML parameter estimates can have infinite magnitude. While we could find MML parameter estimates using a conjugate gradient method, for this paper we constrain the parameter values by calculating maximum a posteriori (MAP) parameter estimates using a generalized Newton's method, and use these values to calculate (11).

Using the decomposition in (1) we can calculate the message length of a network with dag structure $S$:

$$I_{BN} = -\log P(S) + \sum_{Y \in \mathbf{V}} I_Y \quad (12)$$

All that is then required is to define the prior distribution over network structures $P(S)$. We follow Wallace et al. (1996) and use a structure prior that is consistent with their causal interpretation. We consider all total orderings for networks of the same arc density to have equal prior probability:

$$P(S) = \frac{O^S}{m!} p^E (1-p)^{[m(m-1)/2]-E} \quad (13)$$

where $O^S$ is the number linear extensions of $S$, $m$ the number of variables, $p$ our prior belief in the presence of each arc and $E$ the number of arcs in the network.

## 5 MODEL SELECTION

Equation (12) allows us to compare traditional BNs with FOM networks — which we term first-order networks (FONs) — by substituting the relevant local calculation for $I_Y$. We can also use MML to select between conditional probability models on a node-by-node basis. Evaluating both the full conditional model at a node with $I_Y^\Theta$ and the FOM at that node with $I_Y^\Phi$ allows us to "transmit" the node using the model giving the shorter message. To inform the receiver which model we are using requires us to define a prior over local models. For this paper we simply assume that the traditional and first-order models are equally likely and encode the choice with one bit per node. Because one bit of message length is equivalent to a factor of 2 in posterior, we do not perform this local model selection when both models have the same expressive power — i.e., with 0 or 1 parent. Instead we use the traditional model to encode such variables. We term a network using this selection mechanism a dual network. This technique is equally valid for selecting from larger sets of conditional probability models.

As our interest lies in causal discovery, we must also define a search algorithm over the space of causal structures. Proposed techniques range from greedy search (Wallace et al., 1996) to genetic algorithms (Neil and Korb, 1998). For this paper we use a stochastic sampling approach similar to that used by Wallace and Korb (forthcoming) for discovering linear causal models for real-valued data.[5] We use a variant of a Metropolis algorithm (Metropolis et al., 1953) for sampling from the posterior distribution of networks, which we term MML-Sampling.

Our algorithm incorporates the notion of MML-equivalence first described in Wallace and Korb (forthcoming). To assign a particular network to an equivalence class we first consider which arcs may be insignificant. An arc is considered significant if its removal results in an increase in message length. We term a network with all insignificant arcs removed a *clean* model. Cleaning a network before assigning it to an equivalence class effectively associates a posterior with the clean model that is the sum of the posterior of all unclean variants of that model. This strategy can be justified by considering that each unclean version of a model may account for only an small portion of the posterior model distribution, but as a group it is possible for their combined posterior probability to be substantial. In consequence one may discount a substantial group of similar models merely because no single model in the group performs outstandingly. For a particular node, our cleaning procedure proceeds through the parent list in sequence and removes those that are insignificant until such time as there are no insignificant parents left, or the last parent has been tested. While simple, this process is biased in two ways: first, it does not examine all subsets of parents; and second, the subsets it does examine depend upon the order in which parents are considered for removal. It has however proved effective in identifying models accurately in preliminary results. We also aggregate statistically equivalent models into the same MML-equivalence class. While this is justified for the full conditional model, there is no guarantee that statistical equivalence holds for the FOM. However, empirical evidence suggests that it may hold, and for this paper we assume that it does.

In summary, MML-Sampling moves through the space of networks ensuring that models are visited with a probability proportional to their posterior probability as estimated by message length.[6] By keeping counts of each MML-equivalence class visited it reports the best classes in order of posterior probability. Each class is summarized by its cleaned network with highest individual posterior. Reporting several good model classes and estimates of their posteriors has the added advantage of allowing us to use model averaging for prediction.

---
[5] A sampling algorithm for discovery of discrete BNs with the full conditional probability model has also been implemented by Wallace.

[6] We note that while networks are cleaned to determine their MML-equivalence class, they remain unchanged in the Metropolis sampling process.

| Database | Cases | Vars. | Arity | Source |
|---|---|---|---|---|
| Zoo | 101 | 17 | 2-7 | UC Irvine |
| ICU | 200 | 17 | 2-3 | CMU Statlib |
| Flare1 | 323 | 13 | 2-7 | UC Irvine |
| Voting | 435 | 17 | 2-3 | UC Irvine |
| Popularity | 478 | 11 | 2-9 | CMU Statlib |
| Nursery | 12960 | 9 | 2-5 | UC Irvine |

Table 1: Summary of datasets used.

## 6  RESULTS

To compare the traditional probability model with the FOM we selected six datasets from machine learning repositories (summarized in table 1). To evaluate the performance of inferred networks we use a typical cross-validation technique and randomly split each dataset into 90% training and 10% test data. We report results averaged over 10 different training/test sets for each database. We separately infer MML-equivalence classes for traditional Bayesian networks (TBN), first-order networks (FON) and dual networks (DN) on each training set using the MML-Sampling algorithm. Rather than reporting results for the highest posterior MML-equivalence class inferred from each run, reported results are posterior weighted averages over the inferred classes. We then average these values over the 10 training sets from each database. In order to keep the complexity of the discovery algorithm practical we limit inferred models to have a maximum of 10 parents per node, and limit the full CPT to contain less than 65000 parameters.

Table 2 compares posterior weighted message lengths and negative log likelihood ($-LL$) on test data for the three types of network. While the message length results indicate the method preferred on posterior probability, the $-LL$ reports predictive accuracy on unseen data and give an indication of model generalizability. Message length results show the first-order probability model improving the posterior probabilities of inferred models in general. For six out of the seven datasets both the inferred FONs and DNs show highly significant ($p \leq 0.005$ on two-tailed paired t-tests) differences in message length when compared to the TBN. In addition, there is only a single significant message length result reported against the dual model, for the Flare1 database using the FON, and in this case there is virtually no difference in $-LL$ between the two. While the smallest $-LL$ results for each database do not always coincide with the model giving shortest message length on average, where there are significant differences in $-LL$, there is also a significant difference in message length. For three out of the six databases, $-LL$ results favoring the dual network also support the use of MML for selecting the appropriate local probability model at each node, with no significant $-LL$ results reported against the DN.

Several databases warrant further discussion. The Nursery database is synthetic, drawn from a hierarchical decision model for ranking applications to nursery schools. It is a complete database, having one instance for every combination of attributes and the correct class (or ranking). The message length results favor the FON and DN by more than 350 nits, and demonstrate that the first-order interactions outweigh any higher-order interactions in this model. In every run, both the FON and DN methods identified the same network structure: that with all non-class attributes as direct parents of the class attribute and no other arcs. This network is prohibitively expensive to describe using a TBN, requiring a table with > 50000 probabilities and was thus never even examined by the TBN sampling. Of note is the difference in message length between the DN and FON, even though they infer the same structure. Recalling that the DN uses the full conditional model when a node has 0 or 1 parent, this difference is caused by the encoding of the 8 non-class attributes with no inferred parents. For this database, encoding these variables using the first-order model takes about 20 nits more than the same variables encoded with the full conditional model despite the fact that both models have equivalent expressive power for these variables. This difference (noted in Neil et al., 1999) is in part due to the parameter priors assumed for each model, and in part because MAP estimates are used to evaluate the first-order model message lengths rather than MML estimates. This effect, however, does not always favor the traditional probability model, and while it is not an error it must be kept in mind when comparing the FON with the other two methods. The effect may also be alleviated by assuming a conjugate prior for the first-order model, but as mentioned, a symmetric conjugate prior is difficult to define.

The Popularity database surveys children of grades 4-6 on factors effecting popularity at school. While some attributes reflect children's age, grade, gender, etc., four of the 11 attributes represent the participants ranking of grades, sports, looks and money in order of their effect on popularity. The relationship between these four variables is clearly non-monotonic, and cannot be accurately represented by the first-order probability model. There are, however, some monotonic interactions in the database, as evidenced by the enormous (> 200 nit) difference between the FON and the TBN message lengths. In all runs on this database the DN outperformed the FON because it correctly identified the non-monotonic dependency described above, emphasizing the success of MML at selecting the ap-propriate local probability model.

| Data | Message Length (nits) | | | -Log Likelihood (Test Data) (nits) | | |
|---|---|---|---|---|---|---|
| | TBN | FON | DN | TBN | FON | DN |
| Zoo | 760.5 | **713.6** † | 714.3 † | 72.3 | **69.5** | 70.6 |
| ICU | **1220.2** † | 1231.4 | 1221.0 † | 128.3 | **127.6** | 129.4 |
| Flare1 | 2209.8 | **2157.1** ‡ | 2160.1 † | 217.2 | **213.9** | 213.9 † |
| Voting | 4155.8 | 4146.5 † | **4141.2** ‡ | 456.3 | **444.4** † | 450.1 † |
| Popular | 3976.1 | 3768.7 † | **3750.2** ‡ | 385.2 | 378.9 † | **376.8** ‡ |
| Nursery | 113185.0 | 112808.4 † | **112785.8** ‡ | 12520.8 | **12507.7** † | 12507.7 † |

Table 2: Posterior weighted message length and negative log likelihood on test data. Results are averaged over 10 runs for each database and compare the traditional Bayesian network (TBN), first-order network (FON) and the dual network (DN). Boldface indicates minimum average result. ‡ indicates two-tailed paired t-test significance to $\alpha = 0.005$ against both competitors. † shows significance against the method with the worst result.

| Data | Number of Arcs | | | # Parameters | | |
|---|---|---|---|---|---|---|
| | TBN | FON | DN | TBN | FON | DN |
| Zoo | 18.9 | 31.5 | 30.5 | 109.4 | 114.5 | 112.2 |
| ICU | 13.5 | 13.3 | 12.5 | 39.9 | 34.6 | 37.2 |
| Flare1 | 9.6 | 13.1 | 17.0 | 111.9 | 136.4 | 150.7 |
| Voting | 25.5 | 35.7 | 34.5 | 188.4 | 162.2 | 171.8 |
| Popular | 11.7 | 12.3 | 14.4 | 331.2 | 144.1 | 157.8 |
| Nursery | 15.2 | 8.0 | 8.0 | 472.3 | 99.0 | 99.0 |

Table 3: Posterior weighted number of inferred arcs and number of parameters.

The Zoo data categorizes animals based on measured attributes. Although the training databases have only 80 cases, all three methods find some causal structure is warranted by the data. Table 3 gives the average posterior weighted number of inferred arcs and number of inferred model parameters. It is noteworthy that for this database, the FON and DN prefer nearly twice the number of arcs as the TBN on average, yet still manage to fit nearly the same number of parameters to the model. This is a clear indication of the brevity of the first-order probability model. The fact that the −LL are similar for all three models indicates that there may be some (relatively costly) non-monotonic interactions that the TBN is identifying. The FON (and to a less extent the DN) may be able to partially model these non-monotonic interactions by including some of a variable's siblings in its parent set. The siblings of a variable are nodes that share one or more of its direct parents. This introduces additional paths between the parents and the variable (via siblings) and allow for an extended interaction mechanism. This may partly account for the the general trend of Table 3, namely that the FON and DN infer more arcs than the TBN. The trend, however, it is not always followed. The Nursery data gives a counter example where the sample is better modeled with a single large parent set that the TBN cannot identify.

In light of the above discussion it is important to consider the relevance of a causal interpretation of inferred networks under the FOM. When there are no latent variables in a measured system, it is generally acknowledged that with sufficient data, discovered BNs can be interpreted causally, at least to the point of statistical equivalence. While TBNs are capable of expressing arbitrary interactions between direct causes of a variable, limiting the representable causal interactions does have an effect on the causal interpretation of the inferred model. Clearly, a FON can be interpreted causally only when all underlying causal interactions are monotonic. As mentioned in the Zoo database discussion above, violation of this assumption may lead to the inclusion of extra connections that better approximate any non-monotonic interactions. The dual network, however, uses the data to choose the appropriate causal interaction model for each variable. With sufficient data, inferred dual networks can then be interpreted causally, even in the presence of non-monotonic interactions. In fact, it is possible for the dual network to discover the underlying causal system in more detail than the TBN because it is able to efficiently represent monotonic interactions where they are warranted.

## 7 CONCLUSION

We have investigated a logit model of causal interaction, and derived an MML metric for a restricted form of this model exhibiting causal independence. Using a Metropolis sampling approach, we compared Bayesian networks using the traditional full conditional distribution at each node with a network incorporating our restricted first-order model and a dual network that uses MML to select the local model on a node-by-node basis. Tests on six datasets taken from machine learning repositories showed that in five out of the six cases the first-order model had a significant impact on the estimated posterior probabilities of inferred models, and in four cases inferred models that were $> e^{40}$ times more likely a posteriori. In several cases the dual network demonstrated an ability to correctly select the most appropriate local model at each node, obtaining competitive and often significantly better predictive scores on unseen test data. These results illustrate the benefits of increased flexibility and expressive power offered by the dual model over the restricted first-order network (FON) and traditional Bayesian network (TBN). In cases where the dual network did not perform quite as well as the TBN or FON the differ-ences were not found to be significant. A comparison of the average number of inferred arcs indicated that the FON and DN have a tendency to identify more arcs than the TBN, perhaps in an attempt to overcome the limitations of the restricted local distribution. We also tracked a discrepancy in the posterior estimates of the FON and dual network to a difference in parameter prior and estimation, which we intend to address in further work. We also hope to investigate the extension of this approach to less restrictive logit models than the first-order model considered here.